\documentclass[12pt, reqno]{amsart}
\usepackage{amsmath, amsthm, amscd, amsfonts, amssymb, graphicx, color}
\usepackage[bookmarksnumbered, colorlinks, plainpages]{hyperref}
\usepackage{graphicx}
 \graphicspath{{./Pictures/}}
 
\makeatletter \oddsidemargin.9375in \evensidemargin \oddsidemargin
\def\authorsaddresses#1{\dedicatory{#1}}
\numberwithin{equation}{section}
\usepackage{fancyhdr}
\begin{document}

\title[Mask R-CNN for Cadastral Mapping]{Automatic Cadastral Boundary Detection of Very High Resolution Images Using Mask R-CNN}

\author[N. Rahimpour]{Neda Rahimpour Anaraki}
\author[A. Azadbakht]{Alireza Azadbakht}
\author[M. Tahmasbi]{Maryam Tahmasbi}
\author[H. Farahani]{Hadi Farahani}
\author[SR. Kheradpishe]{Saeed Reza Kheradpisheh}
\author[A. Javaheri]{Alireza Javaheri}
\authorsaddresses{Math science Department, Shahid Beheshti University, Tehran, Iran.\\
neda.rpa@gmail.com\\
ali.r.azadbakht@gmail.com\\
m\textunderscore tahmasbi@sbu.ac.ir \\
h\textunderscore farahani@sbu.ac.ir \\
s\textunderscore kheradpisheh@sbu.ac.ir\\
alireza.jvh98@gmail.com\\}


\begin{abstract}

\textbf{Background and Objectives:} Recently there is a high demand for accelerating and improving the detection of automatic cadastral mapping. As this problem is in starting point, there are many methods of computer vision and deep learning that had not been considered yet. In this paper we focus on deep learning and also provide three geometric post-processes methods which improve the quality of the work.

\textbf{Methods:} Our framework included two parts which each of them consists of a few phases. Our solution to this problem is using instance segmentation so in the first part we use Mask R-CNN with the backbone of pre-trained ResNet-50 on the ImageNet dataset. In the second phase we do three geometric post-processes methods in the output of first part to get better overall output. Here also we use help of computational geometry to introduce new method for simplifying lines which we call it pocket-based simplification algorithm.

\textbf{Results:} For evaluating the quality of our solution, we use popular formulas in this field which are recall, precision and F-score. The highest recall we gain is 95\% which also maintain high Precision of 72\%. This resulted to F-score of 82\%.

\textbf{Conclusion:} Implementing instance segmentation using Mask R-CNN with some geometric post-processes to it’s output give us promising results for this field. Also results show that pocket-based simplification algorithms work better for simplifying lines than Douglas-Puecker algorithm. 

\end{abstract}

\maketitle

\section{Introduction}
Cadasters, which record the physical location and ownership of the real properties, are the basis of land administration systems \cite{luo2017investigating}. Nowadays, cadastral mapping has received considerable critical attention. An effective cadastral system formalizes private property rights, which is very important to promote agricultural productivity, secure effective land market, reduce poverty and support national development in the broadest sense \cite{williamson1997justification}. Also recording land rights provides land owners tenure security, a sustainable livelihood and increases financial opportunities \cite{enemark2014fit}.

However, Estimates suggest that about 75\% of the world population does not have access to a formal system to register and safeguard their land rights. Establishing a complete land cadastre and keeping it up-to-date is a contemporary challenge for many developing and developed countries, respectively \cite{enemark2014fit},\cite{luo2017quantifying}. This lack of recorded land rights increases insecure land tenure and fosters existence-threatening conflicts, particularly in developing countries. Recording land rights spatially, i.e., cadastral mapping, is considered the most expensive part of a land administration system \cite{williamson2010land}.

Traditional field surveying approaches to record land parcels are often claimed to be time-consuming, costly and labor intensive. Therefore, there is a clear need for innovative tools to speed up this process \cite{xia2019deep}. Earth observation satellites provide very high resolution (VHR) images. Unmanned aerial vehicle (UAV) images also available in different areas. According to the Union of Concerned Scientists (UCS) there were 971 EO satellites in orbit on the 30th April 2021. For context, when they did a similar report with the end of April 2018 there were only 684 satellites, so there has been a 41.95\% increase over the three years. Whilst this is significant growth, in January 2014 there were only 192 EO satellites according to the UCS database which means over 400\% growth in seven years \cite{pixalytics2012}.
 
Since the availability of VHR and UAV images, remote sensing has been used for mapping cadastral boundaries instead of field surveying, and is advocated by fit-for-purpose (FFP) land administration \cite{enemark2014fit}. In these images, cadastral visible boundaries are often marked by physical objects, such as roads, building walls, fences, water drainages, ditches, rivers, clusters of stones, strips of uncultivated land, etc. \cite{luo2017investigating}. Such boundaries are visible in remotely sensed images and bear the potential to be automatically extracted through image processing algorithms, hence avoiding huge fieldwork surveying tasks \cite{xia2019deep}. 

Convolutional Neural Networks (CNNs) are useful tools in computer vision and image processing tasks. Recent studies indicate that deep learning methods such as CNNs are highly effective for the extraction of higher-level representations needed for detection or classification from raw input \cite{zhu2017deep}, which brings in new opportunities in cadastral boundary detection. Traditional CNNs are usually made up of two main components, namely convolutional layers for extracting spatial-contextual features and fully connected feedforward networks for learning the classification rule \cite{bergado2016deep}. CNNs follow a supervised learning algorithm. Large amounts of labeled examples are needed to train the network to minimize the cost function which measures the error between the output scores and the desired scores \cite{lecun2015deep}.

In deep learning, there are two approaches to train a CNN: From scratch or via transfer learning \cite{garcia2018behavior}. When trained from scratch, all features are learned from data to be provided, which demands large amounts of data and comes with a higher risk of overfitting. An over-fitted network can make accurate predictions for a certain dataset, but fails to generalize its learning capacity for another dataset. With transfer learning, part of the features are learned from a different, typically large dataset. These low-level features are more general and abstract. The network has proven excellence for a specific application. Its core architecture is kept and applied to a new application. Only the last convolution block is trained on specific data of the new application, resulting in specialized high-level features. Transfer learning requires learning fewer features, and thus fewer data \cite{crommelinck2019application}.

In this paper we chose Mask R-CNN with the backbone of pre-trained ResNet-50 on the ImageNet dataset to do instance segmentation as our solution to finding cadastral boundary. We believe this is the first time that instance segmentation being used for cadastral mapping. This method focused on finding boundary of each individual farms and transfer learning for pre-trained ResNet-50 was used. Also we introduce new method for simplifying lines (extracted boundary) which we call it Pocket-based simplification algorithm that will be proven works better than Douglas-Peucker algorithm \cite{crommelinck2016review}, \cite{GRASS_GIS_software}.

In next section we have literature review. then in following section, first we introduce our method which is Mask R-CNN, then as second phase we do some post-processing on the output of network. In section experimental study we talk about training data and the system whole process executed on. Section evaluation focus on investigating how well our presented method works and in the last section we have final conclusion.

\section{Literature Review}
In this section we review some works in this field and talk more about image processing’s different methods to solve problems.
\subsection{Cadastral Mapping}

Some methods for cadastral mapping are based on image segmentation and edge detection. In Drăguţ et al. \cite{druaguct2014automated} they introduce a new automated approach which called Multi-Resolution Segmentation (MRS) to parameterising multi-scale image segmentation of multiple layers. This approach relies on the potential of the local variance to detect scale transitions in geospatial data. Classical edge detection aims to detect sharp changes in image brightness through local measurements, including first-order (e.g., Prewitt or Sobel) and second-order (e.g., Laplacian or Gaussian) derivative-based detection \cite{li2015survey}.

In Crommelinck et al. \cite{crommelinck2017contour} they try to transfer computer vision techniques to remotely sensed UAV images and UAV-based cadastral mapping. They have three steps for finding cadastral boundaries. First is image preprocessing in which the UAV orthoimage was first resampled to lower resolutions and patched. Second, boundary delineation which Globalized Probability of Boundary (gPb) contour detection, a state-of-the-art computer vision method was applied to each patch. This resulted in contour maps containing probabilities for contours per pixel. Third, image postprocessing which all patches belonging to the same image were merged to one contour map and one binary boundary map, which was then vectorized.

There are many studies using CNN tools in cadastral mapping. In Crommelinck et al. \cite{crommelinck2019application} they have a three-steps workflow. First, image segmentation to extract visible object outlines. Second, boundary classification to predict boundary likelihoods for extracted segment lines, and third, interactive delineation to connect these lines based on the predicted boundary likelihood. Image segmentation is based on Multiresolution Combinatorial Grouping (MCG), which delivers closed contours, capturing the outlines of visible objects. Boundary classification is applied to the post-processed MCG lines. They investigate two machine learning approaches to derive the boundary likelihood per MCG line: Random Forest (RF) and pre-trained VGG19 via transfer learning. Interactive delineation supports the creation of final cadastral boundaries by different functions as a plugin in QGIS. All of their training images are UAV.

In Fetai et al. \cite{fetai2019extraction} they also use UAV images. Their workflow consists of three steps. First, image pre-processing. Second, boundary detection and extraction. And for the final step, data post-processing. The first step is resampling the UAV orthoimage. For second step, the UAV orthoimage was resampled from 2 cm to lower spatial resolutions. At the end the ENVI feature extraction module \cite{wang2015road}, \cite{poursanidis2015landsat} was applied to each down-sampled UAV orthoimage. 

Xia et al. \cite{xia2019deep} apply deep Fully Convolutional Networks (FCNs) for cadastral boundary detection based on UAV images acquired over one urban and one semi-urban area. They addressed boundary detection as a supervised pixel-wise image classification problem to discriminate between boundary and non-boundary pixels. The network used in this research is modified from the FCN with dilated kernel (FCN-DK) as described in \cite{persello2017deep}.

\subsection{Preliminaries of Image Processing}

\begin{figure}[h]t
\centering
\includegraphics[scale=0.5]{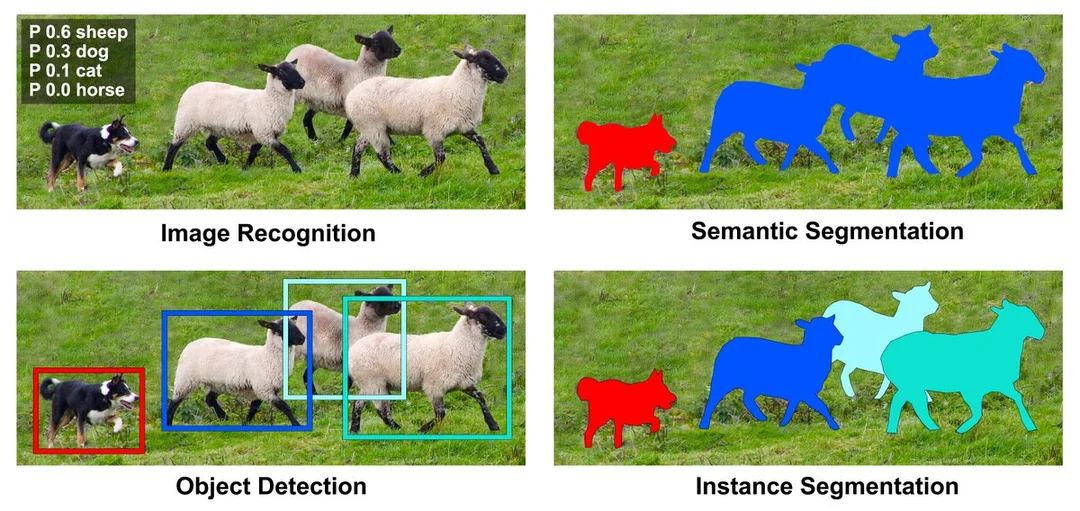}
\caption{Four different types of image processing problems} \label{fig:types}
\end{figure}

Image processing concerns four types of problems that different kind of CNN’s could solve them (Figure \ref{fig:types}):

1. Classification: The input is an image and the problem is finding a class that this image belongs to.

2. Object detection: Here we identify and locate objects in an image by drawing bounding box around each object and determining which class each object belongs to.

3. Semantic segmentation: This problem associates every pixel of an image with a class label by giving specific color to each pixel. It treats multiple objects of the same class as a single entity.

4. Instance segmentation:  It is similar to semantic segmentation but also identifies individual objects within same class.
 
Our solution is based on instance segmentation that has come to be one of the relatively important, complex and challenging areas in computer vision research. Girshick et al. \cite{girshick2014rich} were one of the first to explore CNNs for instance segmentation \cite{liu2020deep}. They developed the R-CNN technique which integrated AlexNet \cite{krizhevsky2012imagenet} along with a region proposal using the selective search technique \cite{van2011segmentation}. But it has some drawbacks like having slow and difficult training. So Girshik \cite{girshick2015fast} by introducing Fast R-CNN, addressed some of the issues of R-CNN, and subsequently improved its object detection ability \cite{hafiz2020survey}.

In Zagoruyko et al. \cite{zagoruyko2016multipath} three modifications have been done to the standard Fast R-CNN model and MultiPath Network introduced. First, skip connections have been incorporated which give the object detector access to features of different network layers. Second, a foveal element has been added for exploitation of context of objects at different resolutions. And finally, a loss function of integral nature has been added \cite{hafiz2020survey}.

In spite of the fact that Fast R-CNN led to significant speed up in detection, it still relied on external region proposals, for which computation was the speed bottleneck in Fast R-CNN. So Faster R-CNN model was proposed by Ren et al. \cite{ren2015faster} which had a Region Proposal Network (RPN) for generation of region proposals and this was efficient and accurate. The same backbone network was used, taking features from the last shared convolutional layer in order to accomplish region proposal by RPN and region classification by Fast R-CNN.

There are other kinds of networks that use for instance segmentation problems that you can read about them in \cite{hafiz2020survey}.

\section{Method}
In this section we present our approach that consists of two main parts: 1. Deep convolutional network that detects boundaries and 2. A geometric post-process that simplify the boundary and cleans up the map.

\subsection{Mask RCNN}

Here we propose an effective pipeline for detecting the borderline of two fields in satellite images. The core module in this pipeline is a Mask RCNN model \cite{he2017mask}. This model is responsible for detecting each field in small input images of 400*400. The RCNN is a two-stage detection algorithm. The first stage identifies a subset of regions in an image that might contain an object, i.e., a field. The second stage classifies the object in each region and returns a probability map of presenting an object, i.e., a field over pixels.

The input images were raw satellite images obtained from Google Maps without additional labels or external data. In order to find borders of fields in a supervised manner, we manually generated labels for some of the input images. With the LabelMe \cite{wada2018labelme} library, we carefully generate a mask for each input image marking each agricultural field area with a unique color. This way, we could train an instance segmentation model to detect each field.

In order to train an instance segmentation model, we chose the Mask RCNN model with the backbone of pre-trained ResNet-50 \cite{he2016deep} on the Imagenet dataset \cite{deng2009imagenet}. Finetuning a pre-trained model would give better results in our experiments. Choosing the input size of this model is a tricky challenge. Small input images would cause the detected object to be more fragmented, large input images cause the model to ignore small features, and small fields would be ignored and considered noise. We chose an input size of 400*400 so that models would detect enough small features and fields and output fields would be too fragmented, and the computational cost of training the model would be reasonable.

A dataset of 400*400 patches was extracted from input images, and Mask RCNN models were trained on them. The model Tries to detect each field as an individual object, and for each object, it returns a probability mask stating whether a pixel is in a field or not. In the challenge, we do not need each field alone, but we want some sense of the borders, so if we could only identify the interior area of each field, we can extract the borders. In order to simplify the overall solution, we apply a max pool on all the detected object masks.

Our solution for this challenge consists of five steps, patchify, Mask RCNN model, unpatchify, edge and border detector, and border refiner. In inference time, we slide a window of 400*400 with steps of 200 pixels on the input image. In this way, the patches overlap. This step is called patchify. each patch goes into the Mask RCNN model, and the resulting masks aggregate together and create the final large output. This step is called unpatchify. On the overlapping parts of patches, we apply many methods. We can apply max pooling, averaging, summation, and harmonic mean depending on the desired output.

Till the unpatchify step, we obtain a large mask of the input model where values close to one are more likely to be field. In order to extract borders from masks, we apply Otsu’s binarizing method \cite{yang1994adaptive} to binarize probability masks adaptively. After binarization, the canny edge detection \cite{canny1986computational} algorithm applies to the output, and the resulting image contains only edges and borders of fields.

\subsection{Geometric Post-Process}

After getting clean output from last part which is binary png image, we need to transform output image to a shapefile so that it could be used in geographic applications like QGIS, Arc GIS, etc. Therefor we extract polygons from the output image and after vectorizing it, shapefile is being created. But three post-process steps should be done at these polygons to get better output.

Before explore these three steps, we need to introduce “pockets” that used in our presented technique of simplifying lines which came out of a concept in computational geometry called convex hull.

The definition of convex hull in an Euclidean space is as follows: A polygon $C$ is said to be convex if for any points $a, b \in C$, every point on the straight-line segment joining them is also in $C$. The convex hull of a polygon $P$ (noted by $CH(P)$) in Euclidean space is the smallest convex polygon enclosing $P$, yielding a polygon connecting the outermost points in the $P$ and all whose inner angles are less than $180$ degrees.

\begin{figure}[h]
\centering
\includegraphics[scale=0.8]{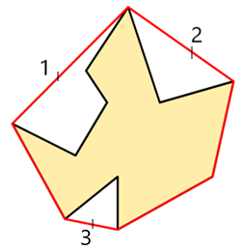}
\caption{Black polygon and it’s convex hull in red lines. Those three marked edges called pockets.} \label{fig:convexhull}
\end{figure}

After making convex hull of a polygon $P$, each edge that belongs to $CH(P)$ but not count as an edge of polygon $P$, called pocket. In Figure \ref{fig:convexhull}, three edges that marked and numberized are pockets. Now let’s get back to three post process steps we need in this section.
 
First, since there are some polygons with area less than a real farm, we delete these polygons due to each image’s ground sample distance (GSD). Second, network detect some polygons inside another polygon. We delete this kind of polygons so that no polygon contains any smaller polygon inside. Third, we need to simplify detected boundaries (lines). For this, there is a popular approach which is Douglas-Peucker Algorithm that we applied to our shapefile so that zigzagged lines become simplify and some over-segmentation lines combine together to form fewer line segments. But we also make a new method for simplifying lines which in evaluation section we will show it works better than Douglas-Peucker Algorithm and we call it Pocket-based simplification algorithm. 
 
 \begin{figure}[h]
\centering
\includegraphics[scale=1]{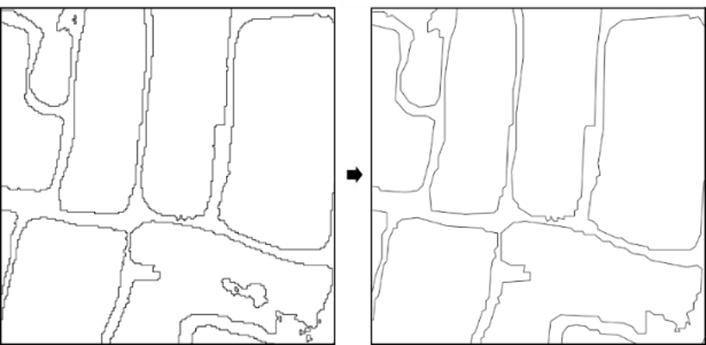}
\caption{Right picture is raw output and left one is output after applying all three steps of geometric post process. (Simplified by Douglas-Peucker algorithm)} \label{fig:whysimp}
\end{figure}

\begin{figure}[h]
\centering
\includegraphics[scale=0.5]{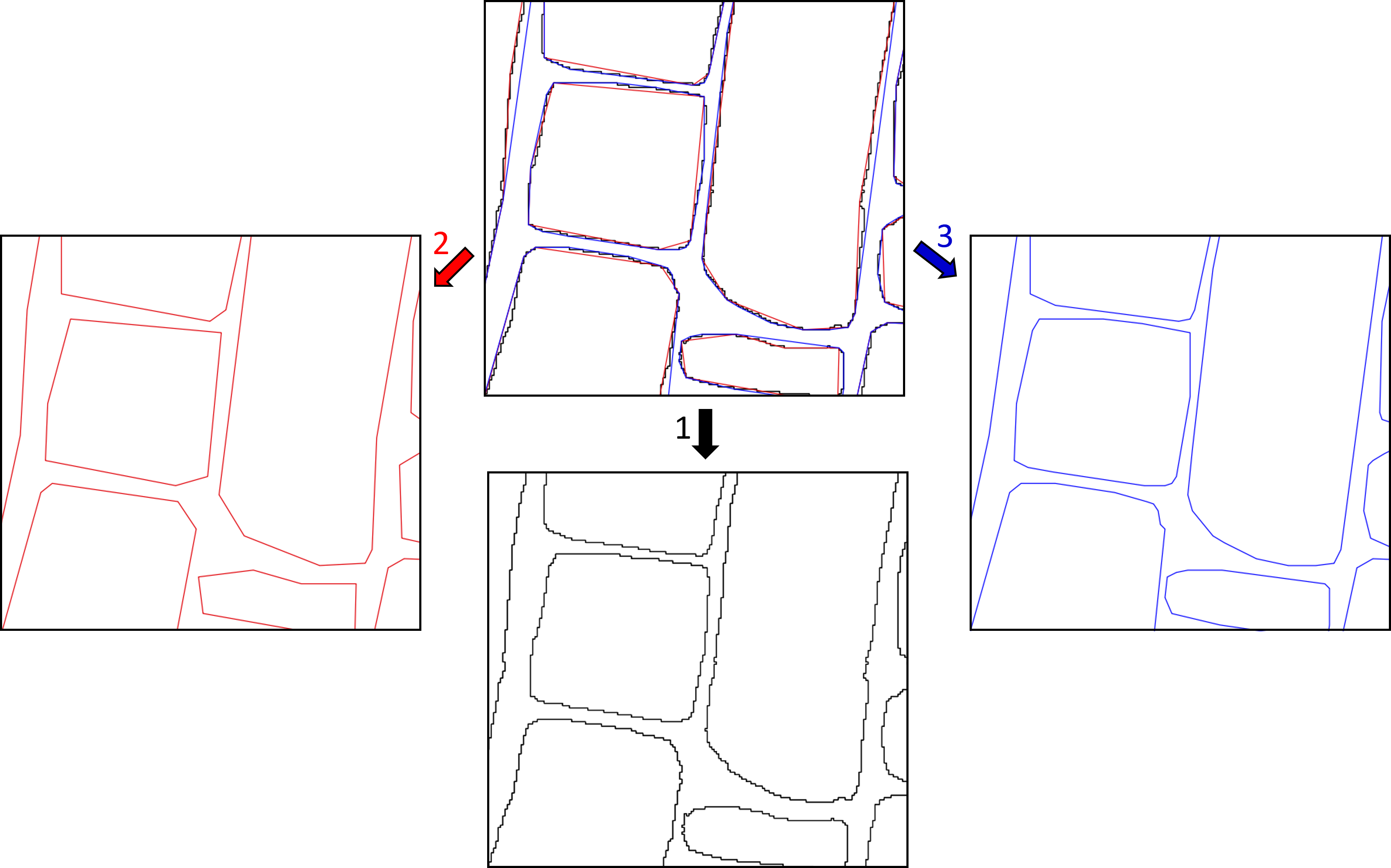}
\caption{Picture of boundary without simplification (1-black), simplified by Douglas-Peucker algorithm (2-red) and simplified by Pocket-based algorithm (3-blue).} \label{fig:diffsimp}
\end{figure}

In this method first we obtain convex hull of the polygon $P$. Then we find all pockets of $P$. Now we calculate two distances for each pocket: First is length of the pocket (noted by dist), and second is summation on the length of edges of $P$ between two end points of that pocket (noted by $d$). Then with certain threshold (noted by $t$), if $d<t \times dist$, all vertices between end points of pocket are being deleted and this pocket adds as an edge of polygon $P'$. Else, we don’t consider this pocket and all edges in $P$ between end points of pocket add as edges of polygon $P'$. At the end polygon $P'$ which is simplified of polygon $P$ added to shapefile instead of $P$ itself. This method is being executed on all of the extracted polygons thus the final simplified output becomes ready. Figure \ref{fig:whysimp} shows how these three geometric postprocess affect raw shapefile. Figure \ref{fig:diffsimp} shows differences between raw output, Douglas-Peucker algorithm and Pocket based simplification algorithm.

\section{Experimental Study}
In this section we present the data for training and how we prepare them to become ready for input of Mask R-CNN and the system whole process executed on.

\subsection{Training Data}

\begin{figure}[h]
\centering
\includegraphics[scale=0.5]{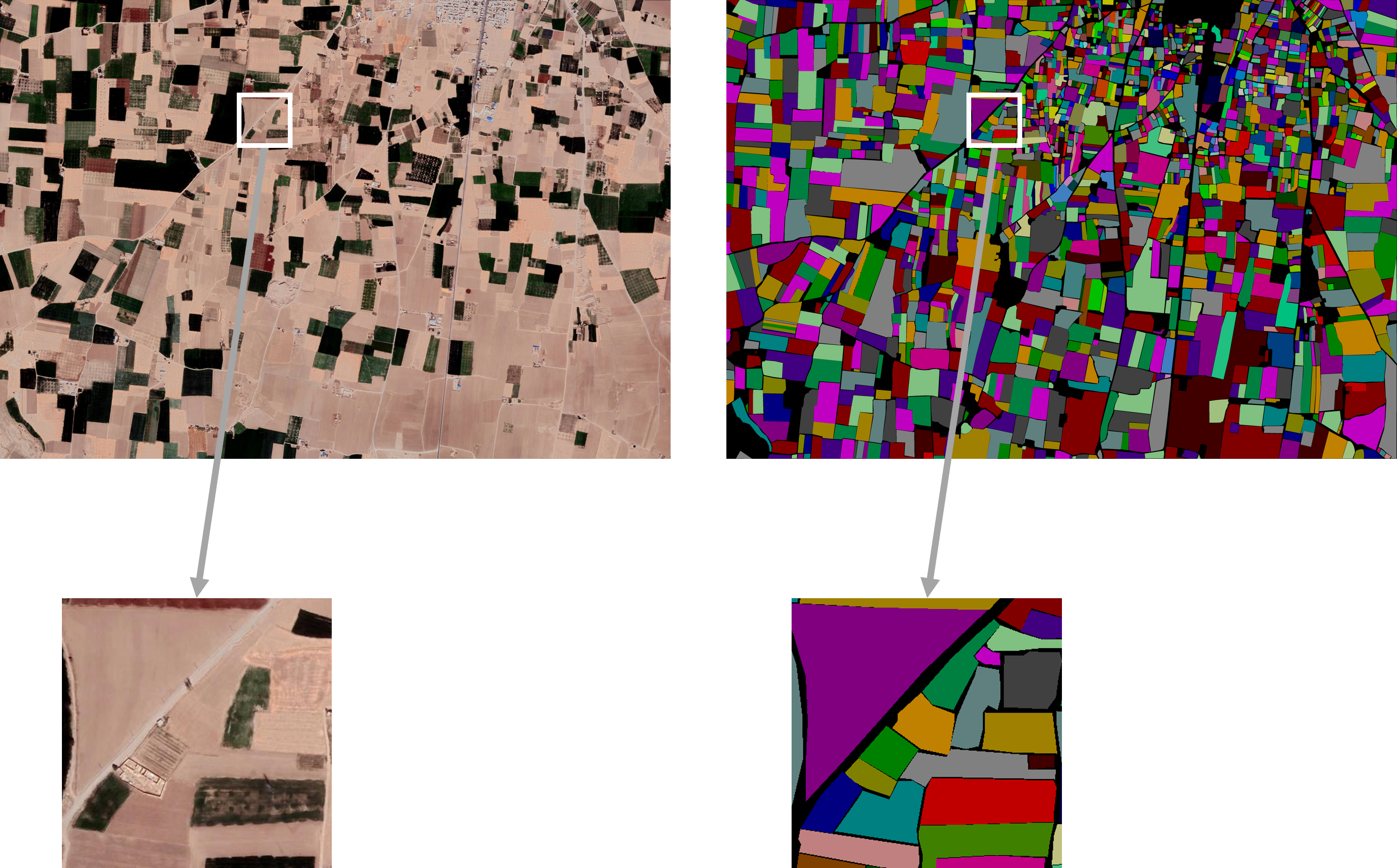}
\caption{Result of a training image with it’s labeled mask.} \label{fig:imgmask}
\end{figure}

We use three images for training the network. Aerial image, Ortho image and FameninIrrigated image which are 4963 × 2819, 3999 × 3999 and 5520 × 3776 pixels respectively. As we are supposed to solve instance segmentation problem, we need to make masks for each image so that the goal achieved. Masks created by ourselves since no dataset online would satisfy our needs. To create masks for training data, we use LabelMe, a free graphical image annotation tool written in Python and use Qt for its graphical interface. For the final steps, Each image patches into tiles of 400  × 400 pixels which results in almost 300 tiles for training data. Figure \ref{fig:imgmask} shows the result of a field together with its labelled mask. The only important thing in masks is that no farms sharing same boundary get same color. So Non-neighbor farms are free to get same colors.

\subsection{System Config}

The whole process of this study implemented in Python. All experiments are conducted on Google Colaboratory having a NVIDIA Tesla T4 GPU with 12 GB RAM. Training time in this system almost take 2 hours.

\section{Evaluation}
In this section we present how well our detected boundary works.

We calculate all accuracy assessments on one of the pictures from Famenin, which is 3432 × 3621 pixels. This picture wasn’t used for training and only used for testing. While validating the predicted results with reference data, a certain tolerance is often used in cadastral mapping \cite{wang2015road}. We consider both 5 and 6 pixels buffer for reference boundary which means original reference boundary has width of 1 pixel while in evaluation we consider reference boundary thick by 5 or 6 pixels. 

The measures we use to evaluate our work are common ones in this field which are precision, recall and F-score. Precision represents whether the assigned boundary is valid. Recall shows the ability of network to find all the boundaries and F-score is harmonic mean between precision and recall which is a good overall measurement for final evaluation. After overlapping our output boundary with buffered reference, we calculate mentioned measures by these formulas:

\begin{equation} \label{formula:precision}
Precision =  \frac{TP}{TP+FP}
\end{equation} 

\begin{equation} \label{formula:recall}
Recall =  BF \times \frac{TP}{TP+FN}
\end{equation} 

\begin{equation} \label{formula:fscore}
F-score =   \frac{2 \times Precision \times Recall}{Precision+Recall}
\end{equation} \\

If we consider binary classification for confusion matrix, class positive stands for pixels labelled as "boundary" and class negative stands for pixels labelled as "non-boundary". Thus we can make confusion matrix and calculate measures. BF in recall formula stands for buffer width the we considered for reference in Famenin test picture and its quantification considered different in each case. The reason behind Recall formula is that while sum of TP + FP indicates the number of total detected boundary pixels, TP + FN indicates the number of total boundaries in the buffered reference, rather than the original reference so we need to divide the sum of TP and FN by BF to get the number of total boundaries in the original reference which its width is one pixel \cite{xia2019deep}.

\begin{table}[h]
\centering
\caption{Accuracy comparison with respect to buffer width and simplification method.}\label{tab:result}
\begin{tabular}{|c|c|c|c|c|}
 \hline
 Buffer width & Method & Precision	& Recall & F-Score  \\
 \hline
 5 pixel & Douglas-Peucker	& 60 & 85 & 70  \\
  \hline
 5 pixel & Pocket-based & 67 & 87 & 76  \\
  \hline
 6 pixel & Douglas-Peucker & 66 & 95 & 78  \\
 \hline
 6 pixel & Pocket-based & 72 & 95 & 82 \\
\hline 
\end{tabular}
\end{table}

As we can see in Table \ref{tab:result}, it is obvious that when we make width of buffer ticker by 1 pixel, both precision and recall become better as more detected boundaries falls into reference boundary buffer. Recall of 95\% in 6 pixels buffer shows Mask R-CNN finds almost all boundaries of the picture. but because of precision less than 80\% , we should note that we detect some boundaries that are not in reference, so they are extra.

If we fix width of buffer, then recall in both methods of simplifying lines are almost equal but precision is always better in Pocket-based algorithm, which means more valid boundaries are detected by our method than Douglas-Peucker algorithm.

The important point is that precision achieved by 5 pixel buffer and paper’s method is 67\%, while this precision achieved by Douglas-Peucker algorithm only when width of buffer is 6 pixel. This means how stronger Pocket-based algorithm works in finding valid boundaries in comparison to Douglas-Puecker algorithm.
Based on better precision and almost equal recall on Pocket-based simplification algorithm, our method always gets higher F-score. Thus with these results we can conclude that our simplifying method, aka Pocket-based simplification algorithm, works better in all tested situations.

\section{Conclusion}
In this study, Mask R-CNN model to solve instance segmentation was used to detect automatic cadastral boundary of VHR images. This Mask R-CNN model is based on transfer learning so it has the backbone of pre-trained ResNet-50 on the ImageNet dataset. We believe this is first time that instance segmentation is being used to solve this problem. Input of network can be in any desired resolution and as the output we obtain a large mask of the input image where values close to one considered as land. In order to extract boundary of each land from masks, we apply Otsu's binarizing method to binarize probability masks adaptively. Then canny edge detection algorithm applies to the output, and the resulting image contains only cadastral boundary. As we use instance segmentation, output of network contains individual boundary for each land. Then after making shapefile from binary png output of network, we use three geometric post-process procedures to improve raw output of Mask R-CNN. In first step, polygons with less than specific area removed based on minimum area of farms of that region could have and image’s GSD. Second step is removing all polygons that are placed inside another polygon. The reason behind this process is obvious, no farm lies on another farm. In the final step, two methods used to simplify detected boundaries. Because final output contains lines which are zigzagged and need to be simplified. There is a famous approach for simplifying lines, Douglas-Peucker algorithm, which also used here. But we didn’t stop here and tried to introduce new method for simplifying lines which we proved that works better than Douglas-Peucker algorithm. We call our introduced method Pocket-based simplification algorithm. The reason behind this name is that it based on produced pockets after making convex hull of each polygon. So based on specific threshold, we make decision on choosing between adding pocket or edges between two end points of a polygon, new simplified polygon has been produced. The way Pocket-based simplification algorithm works, gives us better accuracy on precision, but almost as same recall as Douglas-Peucker algorithm. Which means both have almost same ability to find all cadastral boundary, but Pocket-based simplification algorithm detects more valid boundary. Final evaluation is based on F-score which is harmonic mean between precision and recall and it is always better in Pocket-based simplification algorithm than Douglas-Peucker algorithm. 

Instance segmentation on detecting automatic cadastral boundary represents promising results. Thus using other networks that might works better on instance segmentation may lead a better results.


\bibliographystyle{Unsrt}
\bibliography{references}
\end{document}